\documentclass[runningheads,orivec]{llncs}
\usepackage[T1]{fontenc}
\usepackage{graphicx}
\usepackage{booktabs}
\usepackage[misc]{ifsym}
\usepackage{xcolor}
\newcommand{\corr}{(\Letter)}
\usepackage{amsmath}   % Para comandos matemáticos básicos
\usepackage{amssymb}   % Para símbolos matemáticos extras
\usepackage{mwe}
\begin{document}

\title{TS2TabPFN: Time Series Classification and Extrinsic Regression through Feature Extraction and a Tabular Foundation Model}
\toctitle{TS2TabPFN: Time Series Classification and Extrinsic Regression through Feature Extraction and a Tabular Foundation Model}

\titlerunning{TS2TabPFN}
% If the full title of your paper is short enough to also fit in the running head, you can omit the abbreviated paper title here. You can check as follows: if you comment out the \titlerunning line, something will appear in the header of all odd-numbered pages of your PDF from page 3 onward. This something is either the full title (in which case all is well), or the error message "Title Suppressed Due to Excessive Length". If this error message appears, you're going to want to provide an abbreviated title within the \titlerunning command, because if you won't do it, Springer will do it for you.

%N.B.: Author information (both in the \author{} and \authorrunning{} command) should only be present in the Camera-Ready Version of your paper. The version that you initially submit for review, ought to be double-blind. So, when initially submitting your paper, use:
%\author{Author information scrubbed for double-blind reviewing}

\author{Gabriel da Costa Merlin\inst{1} \and Diego Furtado Silva\inst{1} \corr}
\tocauthor{Gabriel da Costa Merlin, Diego Furtado Silva}

% You may leave out the orcidID information, if you want to.
% Use \corr to indicate the corresponding author. Note the spacing around the \corr command. Only one author can be the corresponding author.

%N.B.: comment out the \authorrunning{} command for the double-blind version of your paper submitted for review. Later, if your paper is accepted, use the command for the Camera-Ready Version.
\authorrunning{G. C. Merlin and D. F. Silva}
% If there are two authors, write 'A.L. Benjamin and C.C. Broadus Jr.'
% If there are more than two authors, '[...] et al.' is used.

\institute{University of São Paulo (USP), São Carlos, SP, Brazil
\email{\{gabrielcmerlin,diegofsilva\}@usp.br}}

\maketitle              % typeset the header of the contribution

\begin{abstract}
Time series data are ubiquitous in practical applications, where classification (TSC) and extrinsic regression (TSER) have emerged as essential tasks for obtaining value from temporal sequences. While the literature has seen significant progress through feature-based and deep learning models, existing methods often focus either on the quality of feature extraction or on the intrinsic predictive power of complex architectures applied to raw data. This division creates a gap between the control offered by feature engineering and the automated performance of end-to-end models. This paper proposes TS2TabPFN, a framework that bridges this gap by integrating explicit feature extraction with TabPFN 2.5, a cutting-edge foundation model for tabular data, to leverage its predictive capabilities. Our extensive experimental evaluation demonstrates that TS2TabPFN significantly outperforms state-of-the-art models in TSER tasks with statistical significance, providing a robust and efficient alternative for TSC and surpassing most of the currently best-performing algorithms. These results suggest that combining foundation models with structured features overcomes single-paradigm limitations, establishing a new time series state-of-the-art.

\end{abstract}

\section{Introduction}

Time series data are ubiquitous in applications ranging from healthcare to activity recognition. While literature historically focused on forecasting, Time Series Classification (TSC) and Extrinsic Regression (TSER) are increasingly recognized for their importance. TSC maps a series to a categorical label, such as identifying a species from audio~\cite{dau2019ucr}. In contrast, TSER aims to predict a continuous scalar value extrinsic to the series, such as glucose levels from physiological signals~\cite{tan2021time}. Despite their distinct outputs, both tasks share the core challenge of learning complex relationships from ordered sequences of observations.

To address these tasks, the literature evolved into distinct categories, including feature-based, convolution-based, and deep learning. Deep models like InceptionTime~\cite{ismail2020inceptiontime} achieve results competitive with costly methods by learning features directly from raw data~\cite{middlehurst2024bake}. While offering high predictive power, these
architectures often lack interpretability. Conversely, feature-based algorithms rely on explicit properties from toolkits like tsfresh~\cite{christ2018time} or catch22~\cite{lubba2019catch22}, providing better control, allowing simpler models to achieve competitive results. 

A third prominent category involves convolution-based techniques, most notably the ROCKET (RandOm Convolutional KErnel Transform)~\cite{dempster2020rocket} family. These methods utilize large numbers of random, unadjusted kernels to extract simple features, prioritizing computational efficiency over complex architectural tuning. While the standard ROCKET-based methods employ linear classifiers or regressors, there is no clear evidence in the literature that more elaborate discriminators consistently achieve superior efficacy. 

Alternatively, meta-ensemble approaches such as HIVE-COTE 2.0 (HC2) \cite{middlehurst2021hive} have demonstrated superior TSC performance by integrating diverse inductive biases within a hierarchical ensemble approach. However, this high predictive accuracy imposes a substantial and often challenging computational burden. Each constituent HC2 module operates as an expensive ensemble, and the complex decision-weighting mechanism requires intensive optimization tuning on the training data. Consequently, such methods can become entirely prohibitive due to their extreme time requirements, especially when processing large datasets.

\begin{figure}
    \centering
    \includegraphics[width=1\linewidth]{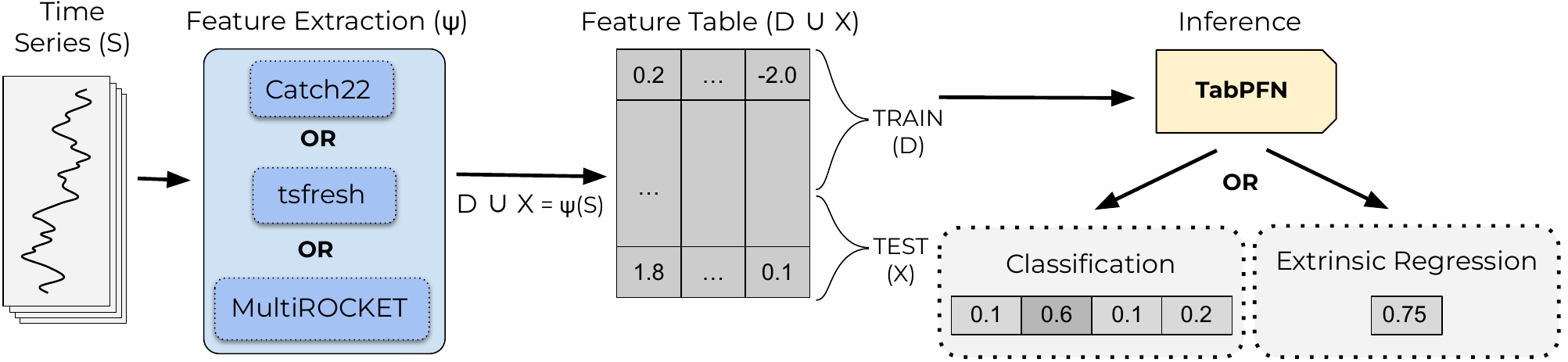}
    \caption{TS2TabPFN framework pipeline. Raw dataset $S$ is processed by a modular function $\psi$, transforming sequences into fixed-dimensional vectors ($d$ for training, $x$ for test). These features form a joint table ($D \cup X$) by concatenating labeled context $D$ and unlabeled queries $X$. Finally, the pre-trained TabPFN model processes this input to generate predictions for Classification or Extrinsic Regression tasks.}
    \label{fig:pipeline}
\end{figure}

We propose TS2TabPFN, a two-step framework for TSC and TSER that combines explicit feature extraction with modern foundation models to mitigate existing paradigms' drawbacks. As shown in Figure~\ref{fig:pipeline}, the first step extracts features using tsfresh, catch22, or the ROCKET family. These are then processed by TabPFN 2.5~\cite{grinsztajn2025tabpfn}, a tabular foundation model pre-trained on synthetic data and optimized for datasets with up to 50,000 samples and 2,000 features, though these are recommended rather than strict limits. Individual forward passes are restricted to 500 features, with the model automatically employing a feature-subsampling ensemble when this threshold is exceeded, a behavior accommodated by our framework, particularly with exhaustive extractors such as tsfresh and MultiROCKET, which often produce high-dimensional representations.

Through extensive experiments, we demonstrate that TS2TabPFN significantly outperforms state-of-the-art models on TSER, with statistical significance. For TSC, our model serves as a high-quality alternative with relatively low computational cost compared to other feature-based and deep learning ensembles. Finally, we acknowledge that TS2TabPFN inherits certain constraints from its underlying foundation model, such as fixed limits on sample size and feature dimensionality due to Transformer memory complexity. While these constraints led to the exclusion of datasets that exceeded available VRAM, the framework remains highly effective for a vast majority of the standard benchmark archive, offering a predictable and efficient workflow within these operational boundaries.

The primary contributions of this work are summarized as follows:
\begin{itemize}
    \item \textbf{Novel Integration:} We present a large-scale empirical study on the application of tabular foundation models to time series, demonstrating that ICL-based architectures scale effectively for TSER. By decoupling feature extraction from the predictive head, we bridge the gap between structured feature engineering and the automated reasoning of foundation models.
    \item \textbf{Extensive Evaluation:} We provide a comprehensive experimental analysis across 189 datasets, demonstrating that TS2TabPFN achieves state-of-the-art performance in TSER and competitive results in TSC.
    \item \textbf{Computational Efficiency:} We show that our training-free, in-context inference approach provides speedups of up to two orders of magnitude compared to established state-of-the-art methods like HC2 and DrCIF.
    \item \textbf{Operational Characterization:} We identify and discuss the practical boundaries of applying foundation models to time series, providing a transparent assessment of how hardware constraints and model architecture influence dataset selection in large-scale benchmarks.
\end{itemize}

\section{Background and Related Work}

In this section, we introduce the necessary definitions and notations, and present the main publications related to our proposal. We begin by formally defining the type of data we address. A time series $s$ with length $T$ is a sequence of $T$ ordered values $s = (s_1, s_2, \dots, s_T)$, $s_i \in \mathbb{R}^{C}$, such that $C$ represents the number of channels of that time series, where $C \in \mathbb{N}^{+}$. Consequently, a time series is classified as univariate if $C = 1$, and as multivariate if $C > 1$.

This paper approaches two tasks: Time Series Classification (TSC) and Time Series Extrinsic Regression (TSER). In TSC, the objective is to learn a mapping $f: \mathcal{S} \rightarrow \mathcal{Y}$, where $\mathcal{Y} = \{c_1, c_2, \dots, c_k\}$ is a finite set of discrete class labels. In contrast, TSER aims to learn a mapping $g: \mathcal{S} \rightarrow \mathbb{R}$ that outputs a continuous scalar. While both tasks involve learning from temporal sequences, they differ in the nature of their target space. We also note that, although both tasks have real-valued outputs, TSER is distinct from time series forecasting. While forecasting aims to predict future observations within the same sequence, TSER focuses on estimating an external target that is not a future value of the input series.

The proposed TS2TabPFN (detailed in Section~\ref{sec:proposal}) is closely related to existing algorithms for TSC and TSER, particularly those based on feature extraction. The remainder of this section first describes well-stabilized and widely used feature suites that comprise the basis of relevant TSC and TSER algorithms. Subsequently, it discusses state-of-the-art algorithms for these tasks and concludes by explaining the tabular foundation model TabPFN.

\subsection{Feature Extraction for Time Series}\label{sec:features}
\label{sec:background_feat}

The choice of the feature suite is a central component of feature-based TSC and TSER methods \cite{middlehurst2021hive}. Among the most widely used approaches is the Time Series Feature Extraction based on Scalable Hypothesis tests (tsfresh) \cite{christ2018time}, which automatically extracts hundreds of statistical and signal-processing features from time series. To control the resulting high-dimensional feature space, tsfresh applies scalable hypothesis tests to retain only features that are statistically relevant to the target variable. The library provides different extraction profiles, with the efficient configuration being commonly adopted in practice due to its favorable trade-off between computational cost and predictive performance.

The CAnonical time series CHaracteristics (catch22)~\cite{lubba2019catch22} is another significant contribution to the field. Unlike exhaustive libraries, catch22 provides a compact set of 22 features selected from the thousands available in the hctsa toolbox. These features were chosen through a rigorous process to identify a subset that is minimally redundant, computationally efficient, and highly discriminative across a diverse range of 93 classification datasets. Because catch22 is scale-agnostic and avoids the heavy computational overhead of larger libraries, it may be particularly well-suited for scenarios with limited resources.

We also consider ROCKET-based methods as feature extraction approaches, as they transform raw time series into fixed-length feature vectors. The original ROCKET \cite{dempster2020rocket} applies thousands of random convolutional kernels and summarizes their outputs using simple statistics such as the maximum value and the Proportion of Positive Values (PPV). MultiROCKET \cite{tan2022multirocket} extends this idea by incorporating first-order differences and additional pooling operators, increasing feature diversity while maintaining high computational efficiency.

\subsection{Learning Algorithms for Time Series}
\label{subsec:learning_algs}

The simplest, yet effective, algorithm that explicitly leverages feature extraction for TSC or TSER is the tsfresh Pipeline with Rotation Forest (FreshPRINCE)~\cite{middlehurst2022freshprince}. This method consists of a two-step approach, where it represents the time series using the tsfresh library and feeds it into a Rotation Forest classifier or regressor~\cite{bagnall2018rotation}. The Rotation Forest applies Principal Component Analysis (PCA) to random feature subsets and trains decision trees on the transformed space. This operation enables Rotation Forest to handle the multicollinearity and high dimensionality inherent in tsfresh outputs. 

In contrast, the ROCKET family~\cite{dempster2020rocket,tan2022multirocket} (including MiniRocket and MultiROCKET) avoids manual feature engineering by employing a large bank of random convolutional kernels. Similar to FreshPrince, ROCKET is a two-step approach. It starts by extracting simple features from the signal resulting from convolving these kernels with the time series (c.f. Section~\ref{sec:features}). After transforming the data into this high-dimensional feature space, it employs traditional linear models as the final predictive head for both classification and regression.

Choosing between these techniques for a new dataset is difficult. While ROCKET-based methods achieved better results on TSC benchmarks~\cite{middlehurst2024bake}, recent evidence suggests that 
FreshPRINCE often performs better on TSER~\cite{guijo2024unsupervised}.

The Diversified Random Convolutional Interval Forest (DrCIF)~\cite{middlehurst2021hive} proposes extracting features in a more local approach. Unlike global methods, such as FreshPRINCE and ROCKET, that summarize the entire series, DrCIF operates over random intervals. Specifically, it extracts catch22 and summary statistics (mean, standard deviation, and slope) from raw data, periodograms, and auto-regressive of randomly chosen subsequences. By focusing on intervals, DrCIF captures local patterns that may be weakened by global feature extraction. This interval-based approach has proven effective for TSC, and extensive analysis has shown it stands as one of the most accurate methods for TSER~\cite{guijo2024unsupervised}.

As an alternative to explicit feature extraction, deep learning introduced end-to-end neural architectures for TSC and TSER~\cite{mohammadi2024deep}. Baseline models like Fully Convolutional Network and Residual Network use stacked convolutional layers to learn filters directly from raw signals~\cite{wang2017time}. InceptionTime~\cite{ismail2020inceptiontime} refined this via Inception modules to capture patterns at varying scales. However, these models face constraints: to achieve state-of-the-art results, InceptionTime typically requires ensembles of five or more models, increasing computational costs for training and inference, largely due to high variability across training runs. Furthermore, these architectures do not consistently outperform established feature-based methods across diverse benchmarks and experimental settings.

Finally, we note that the most notable TSC technique in recent benchmark assessments is HIVE-COTE 2.0 (HC2)~\cite{middlehurst2021hive}. HC2 is a hierarchical meta-ensemble that combines four distinct modules: DrCIF (interval-based), TDE (Temporal Dictionary Ensemble), STC (Shapelet Transform Classifier), and Arsenal (an ensemble of ROCKET classifiers). By integrating algorithms with different inductive biases, HC2 achieves outstanding robustness across diverse data types. Nevertheless, HC2’s accuracy comes at a prohibitive computational cost. Beyond the expense of running its individual constituent ensembles, HC2 requires a sophisticated weighting mechanism (CAWPE - Cross-validation Accuracy Weighted Probabilistic Ensemble) to estimate the reliability of each module. This process involves intensive internal cross-validation on the training set, making HC2 difficult to scale to large-scale datasets or resource-constrained environments.

\subsection{TabPFN: Tabular Prior-Data Fitted Network}

Unlike traditional supervised machine learning approaches that require training a model from scratch on every new dataset, TabPFN (Tabular Prior-Data Fitted Network)~\cite{hollmann2023tabpfn} operates as a tabular foundation model. It performs direct inference, i.e., it generates predictions in a single step through a pre-trained Transformer architecture. To achieve this, TabPFN is designed as a Prior-Data Fitted Network (PFN) that approximates complex Bayesian reasoning, enabling it to recognize patterns and make highly accurate predictions on new, unseen data without additional training or hyperparameter tuning.

Fundamentally, TabPFN bypasses iterative parameter optimization by framing tabular prediction through the lens of In-Context Learning (ICL). Within this paradigm, a given training dataset $D = \{(d_i, y_i)\}_{i=1}^{n}$ is used strictly as context, rather than as data to update the model's internal weights. Together with an unlabelled query point $x$, it is provided directly to the model as a permutation-invariant, set-valued input. Rather than learning a single optimal mapping from scratch, the pre-trained Transformer performs a single forward pass to approximate the true probability of the target label $\hat{y}$ given the observed evidence. In Bayesian statistics, this comprehensive prediction is known as the Posterior Predictive Distribution (PPD), formally defined as:
\begin{equation}
p(\hat{y} \mid x, D) \propto \int_{\Phi} p(\hat{y} \mid x, \phi)\, p(D \mid \phi)\, p(\phi)\, d\phi.
\end{equation}

\noindent Under the Bayesian supervised learning framework, $\Phi$ represents the encompassing space of all possible hypotheses (i.e., data-generating mechanisms) that map inputs to outputs. Exact Bayesian inference dictates that the prediction for a new query should be obtained by marginalizing over this entire hypothesis space. As formalized in the integral, the final output aggregates the predictions from all hypotheses. Specifically, the prediction made by each hypothesis for the query point, denoted as $p(\hat{y} \mid x, \phi)$, is weighted by both its prior probability $p(\phi)$ and its likelihood $p(D \mid \phi)$ of having generated the observed training data $D$. 

Because calculating this integral analytically is computationally intractable for complex tabular data, TabPFN relies on amortized inference. The Transformer is trained offline on millions of synthetic datasets, each constructed by explicitly sampling a ground-truth mechanism from a mathematically defined prior $p(\phi)$. Consequently, the network learns to implicitly execute this complex Bayesian marginalization, allowing it to rapidly perform highly accurate zero-shot inference on new, previously unseen datasets.

In its latest release, TabPFN 2.5~\cite{grinsztajn2025tabpfn} significantly expanded its operational regime, supporting tabular datasets with up to approximately 50,000 samples and up to 2,000 features as recommended limits, while internally partitioning the feature space into 500-feature subsets during inference. Beyond maintaining competitiveness with state-of-the-art AutoML systems like AutoGluon, TabPFN 2.5 demonstrates superior predictive robustness in regression tasks, outperforming gradient-boosted trees such as XGBoost in the majority of experiments.

\section{TS2TabPFN: Proposed Framework}\label{sec:proposal}

We propose TS2TabPFN, a modular framework that leverages the predictive power of tabular foundation models for time series data. The core direction of the framework is to explicitly decouple the temporal representation from the inferential engine, allowing the model to process complex time series patterns through an ICL paradigm. As illustrated in Figure~\ref{fig:pipeline}, the framework operates in a two-stage pipeline. First, a raw time series $s \in \mathbb{R}^{C \times T}$, where $C$ is the number of channels and $T$ is the sequence length, is transformed into a fixed-dimensional feature vector using a dedicated extraction function $\psi$:
\begin{equation}
    d = \psi(s_{train}),
\end{equation}
\begin{equation}
    x = \psi(s_{test}).
\end{equation}

By applying $\psi$ to every sample in a time series training set, we transform the temporal problem into a tabular format, generating the exact context dataset $D = \{(d_i, y_i)\}_{i=1}^{n}$ required as input context by TabPFN. This tabularization allows us to utilize TabPFN 2.5 as the downstream inferential head.

Unlike traditional models that require gradient-based optimization for new tasks, TS2TabPFN uses ICL by passing the entire featurized training context $D$ along with a set of unlabelled test queries $X = \{x_i\}_{i=1}^n$ in a single forward pass:
\begin{equation}
    \hat{Y} = f_{\theta}(X, D),
\end{equation}
where $f_{\theta}$ represents the frozen weights of TabPFN 2.5 approximating the PPD, and $\hat{Y}$ denotes the final predictions for all test instances. This allows the model to generalize across datasets by identifying patterns in the distribution of the extracted temporal descriptors without any parameter updates.

A significant advantage of this framework is its inherent modular adaptability; the extraction function $\psi$ can be tailored to capture different data characteristics. In this work, we introduce three distinct instantiations:
\begin{itemize}
    \item \textbf{C22-TabPFN}: Extracts 22 canonical features via aeon's~\cite{aeon24jmlr} catch22, covering autocorrelation, distribution spread, and scaling behavior. For multivariate, features are computed per channel and concatenated into one vector.
    
    \item \textbf{tsfresh-TabPFN}: Extracts comprehensive statistical descriptors using the EfficientFCParameters profile from tsfresh to exclude computationally expensive non-linear metrics. Training and test sets are concatenated prior to extraction, and missing values are resolved via the native impute function.
    
    \item \textbf{MultiROCKET-TabPFN}: Leverages multiple pooling operators and random convolutional transformations to map raw series into a highly discriminative feature space using aeon's native MultiROCKET transformer. It applies default configurations to prevent per-dataset hyperparameter tuning
\end{itemize}

By maintaining the TabPFN weights frozen, the framework functions as an efficient inference engine. This eliminates the intensive computational cost of per-dataset training, a common bottleneck in state-of-the-art TSC and TSER algorithms, while maintaining highly competitive performance.

\section{Experimental Methodology}

This section delineates the empirical framework and the corresponding protocol employed to evaluate the performance of TS2TabPFN. We provide a rigorous characterization of the benchmark archives, discuss the technical boundary conditions imposed by memory constraints, and justify the selection of state-of-the-art baselines used for comparative analysis. Furthermore, we describe the standardized evaluation metrics and statistical tests utilized to ensure the validity and reproducibility of our findings across both TSC and TSER tasks.

\subsection{Datasets and Experimental Scope}
\label{subsec:datasets}

To assess the generalization capabilities of TS2TabPFN, we utilized the most prominent benchmarks in the time series community. For TSC, we employed the combined UEA and UCR Archive~\cite{bagnall2018ueamultivariatetimeseries,UCRArchive2018}, comprising $158$ datasets spanning diverse real-world application domains. For TSER, we utilized the TSML Extended Archive~\cite{tsml_archive_maintainers_2024_11236865}, which consists of $63$ regression datasets for rigorous evaluation. Our preprocessing protocol adheres strictly to the standards established by the respective benchmarks. For TSC, datasets natively feature z-score normalization applied to individual sequences as per the UCR archive specifications, whereas TSER tasks retain their original signal scales. Missing values across both archives were handled using linear interpolation where necessary.

A primary architectural requirement of TabPFN is the simultaneous loading of both training and test sets into GPU memory to enable ICL during the forward pass. Due to the $\mathcal{O}(n^2)$ memory complexity inherent in the Transformer-based backbone, datasets characterized by exceptionally long sequences or large sample counts exceeded the available VRAM. Consequently, we refined our experimental scope to a subset of these archives: $134$ datasets for TSC and $55$ for TSER. The specific datasets included and excluded are listed in our GitHub repository\footnote{\url{https://github.com/gabrielcmerlin/TS2TabPFN}}, whereas the complete structural metadata for all benchmark suites remains accessible via the original archive portals\footnote{\url{https://www.cs.ucr.edu/\string~eamonn/time\_series\_data\_2018/}\\\ \url{http://tseregression.org}}. This selection represents the maximal feasible benchmark within our hardware constraints while maintaining a statistically robust and diverse representation of real-world temporal challenges. A comprehensive discussion of these limitations is detailed in Section~\ref{sec:limitations}.

\subsection{Evaluation Protocol and Statistical Validation}

Following the standard evaluation protocol in time series literature~\cite{guijo2024unsupervised,middlehurst2024bake}, we report the average performance across 30 resamples. To ensure a rigorous comparison, we utilized the exact same seeds and splits as the original benchmarks, allowing for direct alignment with results reported in the literature while maintaining identical conditions for the TS2TabPFN framework.

Statistical significance is assessed via Wilcoxon signed-rank tests with Holm's correction ($\alpha$=0.05) for pairwise comparisons. To visualize relative performance across the benchmark, we utilize Critical Difference (CD) diagrams~\cite{demvsar2006statistical}. These diagrams represent the average rank of each model—where lower ranks indicate superior performance—and incorporate post-hoc significance tests to identify cliques. Within the diagram, thick horizontal bars connect models between which there is no statistically significant difference, allowing for a clear identification of state-of-the-art methods while filtering out marginal performance gains.

\subsection{Baseline Models and Benchmarking}
\label{subsec:baselines}

In order to evaluate the competitiveness of TS2TabPFN, we compare our framework against the strongest results reported in the literature, avoiding local re-execution of the baselines to preserve strict alignment with the community's established benchmarks. We chose a set of high-performance models that define the current state-of-the-art in time series analysis. These baselines encompass the different algorithmic categories discussed in Section~\ref{subsec:learning_algs}, providing a comprehensive comparison across different inductive biases.

\begin{itemize}
    \item \textbf{TSC Baselines:} The accuracy is compared against DrCIF, ROCKET, Arsenal, HC1, and HC2, with the latter included as the current gold standard for predictive accuracy on the UCR/UEA archive.
    \item \textbf{TSER Baselines:} Root Mean Squared Error (RMSE) is compared against FreshPrince, InceptionTime, MiniROCKET, MultiROCKET, RotationForest, and DrCIF, the latter being the best for interval-based representation.
\end{itemize}

\subsection{Hardware and Software Environment}

To ensure consistency of runtime measurements and performance benchmarks, all experiments for TS2TabPFN variations were executed on two identical workstations. Each one was equipped with an Intel Core i7-10700F (8 cores, 16 threads), 32 GB of DDR4 RAM, and a GeForce RTX 3060 with 12 GB of VRAM. The software was built on the aeon library for standardized data handling and the official TabPFN implementation\footnote{\url{https://github.com/PriorLabs/tabpfn}} as the core inference engine. 

Adhering to the Prior-Data Fitted Network (PFN) philosophy, no hyperparameter tuning was performed on a per-dataset basis. This strategy ensures that the reported results reflect the model's robust ICL capabilities, providing a transparent assessment of its generalization power across the diverse UCR/UEA and TSML archives without iterative optimization.

\section{Results and Discussion}

We evaluate the proposed TS2TabPFN framework on TSC and TSER tasks. Our evaluation begins by identifying the optimal feature extraction strategy tailored for the TabPFN 2.5 architecture. Subsequently, we benchmark the resulting pipeline against a comprehensive suite of well-established baselines across both tasks. Furthermore, we conduct a rigorous runtime analysis alongside an in-depth assessment of energy consumption, comparing our proposed method directly against the top-performing baseline from each respective task.
 
\subsection{Classification}
\label{subsec:tsc}

\begin{figure}[!htp]
    \centering
    \includegraphics[width=0.8\linewidth]{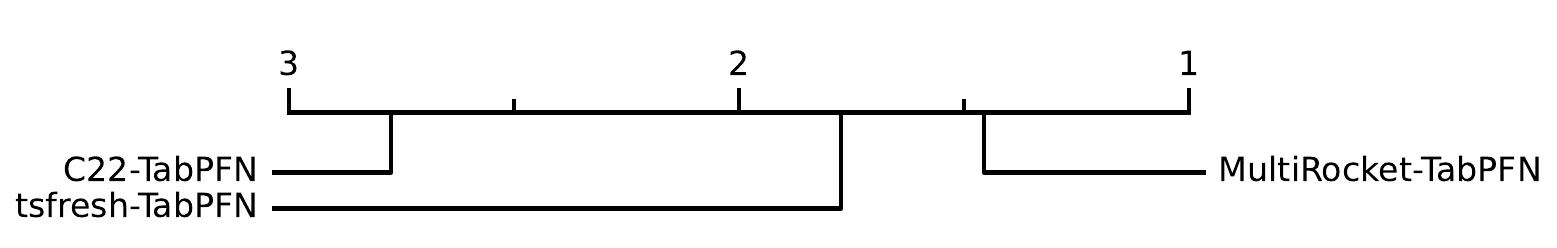}
    \caption{Critical Difference diagram for TSC (Internal Validation) based on pairwise Wilcoxon signed-rank tests of classification accuracy. Closer to 1, the better.}
    \label{fig:tsc_intra}
\end{figure}

As TS2TabPFN integrates feature extraction with a tabular foundation model, the methodology used to derive these features from raw data is a core component of our framework. In this regard, we evaluated three distinct extraction approaches to determine their impact on classification performance. The results of this intra-framework comparison are visualized through the CD diagram in Figure \ref{fig:tsc_intra}. The MultiROCKET-based feature extraction yielded the best performance across the benchmark, achieving statistically significant improvements over the other algorithms tested.

\begin{figure}[!htp]
    \centering
    \includegraphics[width=0.8\linewidth]{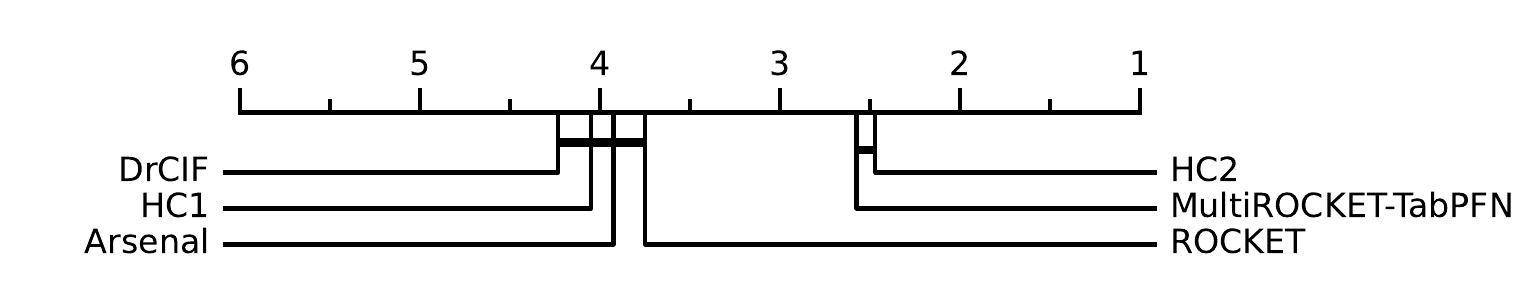}
    \caption{Critical Difference for TSC (Comparative Analysis) based on pairwise Wilcoxon signed-rank tests of classification accuracy. Horizontal black bar indicates groups of models with no statistically significant difference. Closer to 1, the better.}
    \label{fig:tsc_bench}
\end{figure}

Having selected among the assessed feature sets, we now compare our best-performing TS2TabPFN configuration against a selection of top-tier baselines from the literature. The results of this external benchmarking are presented in Figure \ref{fig:tsc_bench}. They indicate that our approach achieves performance parity with the state-of-the-art HC2 and demonstrates statistically superior results over all other evaluated methods across the benchmark datasets.

\begin{figure}[!htp]
    \centering
    \includegraphics[width=1\linewidth]{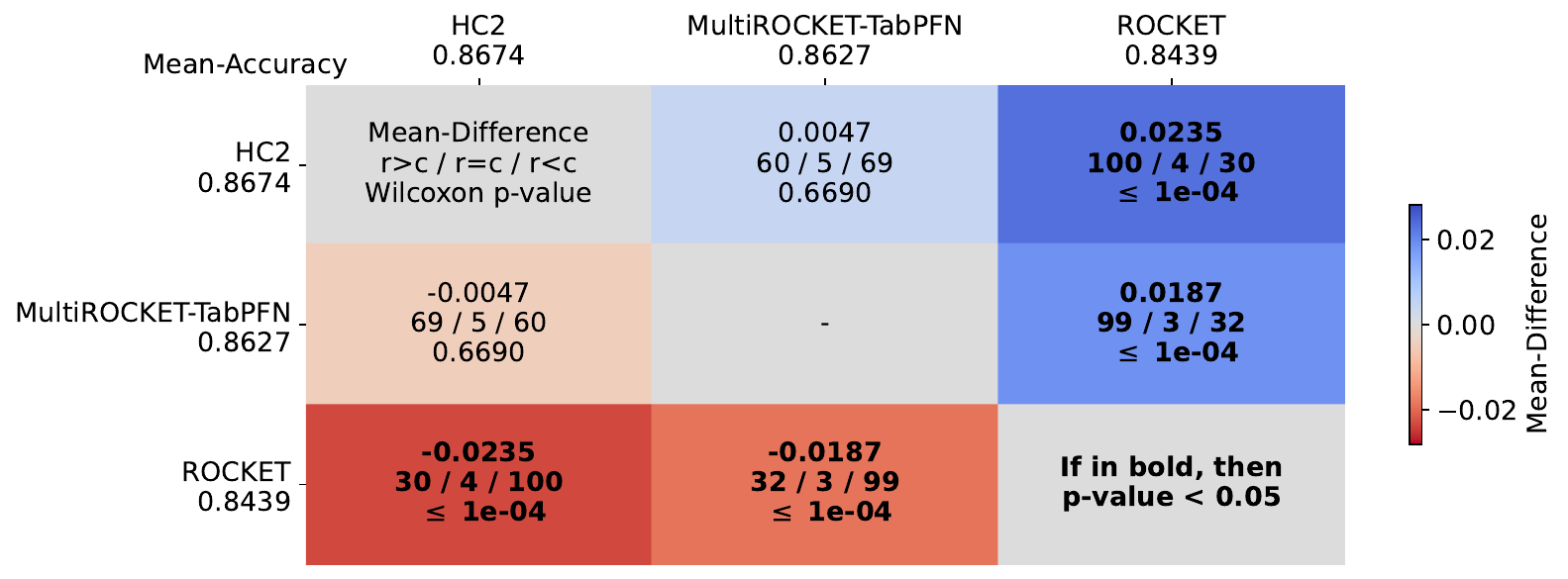}
    \caption{Multiple comparison matrix showing mean differences, win/tie/loss records and Wilcoxon p-values. Bold values indicate statistical significance ($p < 0.05$).}
    \label{fig:tsc_mcm}
\end{figure}

To further compare the top-ranked algorithms, the pairwise Wilcoxon signed-rank matrix (Figure~\ref{fig:tsc_mcm}) reveals that MultiROCKET-TabPFN outperforms HC2 in direct head-to-head wins (69/5/60). However, the mean accuracy of HC2 (0.8674) remains marginally higher than ours (0.8627). This discrepancy, coupled with a non-significant p-value (0.6690), confirms that although our framework wins more frequently, the difference in accuracy across many datasets is not large relative to HC2's. In contrast, against standard ROCKET, our method demonstrates clear statistical dominance ($p \le 1e-04$), securing a 99/3/32 win-tie-loss record.

While previous experiments emphasize the good accuracy of our approach, its practical utility extends to computational efficiency. Crucially, as noted in Section~\ref{subsec:baselines}, baseline metrics were compiled directly from literature, meaning these models were not comprehensively re-executed on our local hardware. Given that full local execution of complex ensembles such as HC2 across all 134 datasets would be computationally prohibitive given our hardware constraints, we instead conducted a targeted runtime analysis. We only locally executed the HC2 baseline implementation alongside MultiROCKET-TabPFN across five representative datasets carefully selected to span a wide spectrum of scales, as detailed in Table~\ref{tab:runtime_tsc}. For a fair comparison, we measure the complete end-to-end operational pipeline of each method: for HC2, this encompasses both training and testing runtimes, whereas for our approach, the recorded runtime reflects the sum of feature extraction, context preprocessing, and inference.

\begin{table}[!htp]
\caption{Computational runtime and speedup factor for the MultiROCKET-TabPFN compared to the HC2 baseline across datasets of varying scales: Small (S), Small-Medium (SM), Medium (M), Medium-Large (ML), and Large (L).}\label{tab:runtime_tsc}
\centering
\setlength{\tabcolsep}{10pt} % Espaçamento generoso entre colunas
\renewcommand{\arraystretch}{1.1} % Respiro vertical
\begin{tabular}{lrrr} % Esquerda para nomes, Direita para números
\toprule
\textbf{Dataset (Scale)} & \textbf{HC2 (s)} & \textbf{MR-TabPFN (s)} & \textbf{Speedup} \\ 
\midrule
BeetleFly (S)      & 156.54    & 68.46    & 2.29$\times$  \\
ACSF1 (SM)         & 1,858.42  & 90.22    & 20.60$\times$ \\
CricketX (M)       & 2,954.87  & 586.90   & 5.03$\times$  \\
CinCECGTorso (ML)  & 11,186.76 & 113.70   & 98.39$\times$ \\
LSST (L)           & 14,151.27 & 1,224.53 & 11.56$\times$ \\ 
\bottomrule
\end{tabular}
\end{table}

The results reveal a significant reduction in training and inference time for MultiROCKET-TabPFN compared to the HC2 baseline. Our framework consistently outperformed HC2 across all scales, with speedup factors ranging from 2.29$\times$ to 98.39$\times$. Notably, in the Medium-Large and Large categories, the framework reduced processing times from several hours to just a few minutes. This gain is most prominent in the CinCECGTorso dataset, where our approach was nearly two orders of magnitude faster than HC2. These findings suggest that our method not only maintains high predictive performance but also offers a more scalable and efficient solution for large-scale time series tasks.

While a clear advantage of the ICL paradigm in our framework is bypassing per-dataset training, evaluating execution speeds strictly at test time isolates pure architectural efficiency. Table~\ref{tab:inference_tsc} presents this inference-only comparison, removing any overhead not related to inference from all baseline metrics. Even under this restricted setup, MultiROCKET-TabPFN maintains an advantage across all scales, achieving speedups from $1.48\times$ to $263.10\times$ on the CinCECGTorso dataset. These results confirm that our efficiency is not merely a byproduct of eliminating training and demonstrate that our inference remains inherently faster than the HC2 baseline. This extreme gap on CinCECGTorso is driven by its asymmetric split (40 train vs. 1,380 test samples), which heavily penalizes HC2 due to its costly instance-by-instance testing transformations.

\begin{table}[!htp]
\caption{Inference runtime and speedup factor comparison between MultiROCKET-TabPFN and HC2 during testing across representative datasets of varying scales: Small (S), Small-Medium (SM), Medium (M), Medium-Large (ML), and Large (L).}\label{tab:inference_tsc}
\centering
\setlength{\tabcolsep}{8pt} % Slightly reduced padding
\renewcommand{\arraystretch}{1.1}
\begin{tabular}{lrrr}
\toprule
\textbf{Dataset (Scale)} & \textbf{HC2 (s)} & \textbf{MR-TabPFN (s)} & \textbf{Speedup} \\ 
\midrule
BeetleFly (S)      & 49.06     & 33.18    & 1.48$\times$  \\
ACSF1 (SM)         & 671.96    & 34.15    & 19.68$\times$ \\
CricketX (M)       & 922.08    & 578.75   & 1.59$\times$  \\
CinCECGTorso (ML)  & 10,613.41 & 40.34    & 263.10$\times$ \\
LSST (L)           & 2,913.92  & 1,192.28 & 2.44$\times$  \\ 
\bottomrule
\end{tabular}
\end{table}

Table~\ref{tab:power_energy_tsc} contrasts operational power rates against total energy consumption across the evaluated scales. Although MultiROCKET-TabPFN exhibits a higher power draw (up to $102.61\text{ W}$) by actively engaging parallel GPU acceleration, the CPU-bound HC2 baseline maintains a lower but constant draw ($\sim 35\text{ W}$) while leaving the graphics hardware completely idle. The reported energy consumption values were obtained using the CodeCarbon library~\cite{lacoste2019quantifying}, while average power rates were computed post-hoc as $P = E_{\text{total}} / t_{\text{total}}$, where $E_{\text{total}}$ and $t_{\text{total}}$ denote, respectively, the total energy consumption and total execution time across all steps of the method. Crucially, this localized hardware intensity is heavily offset by our framework's massive speedups. By avoiding prolonged training cycles and expensive instance-by-instance testing loops, MultiROCKET-TabPFN drastically reduces cumulative energy drawn across all datasets (e.g., from $102.76\text{ Wh}$ to $1.66\text{ Wh}$ on CinCECGTorso), establishing a highly sustainable and eco-efficient deployment footprint.

\begin{table}[!htp]
\caption{Hardware power rate (W) and cumulative energy consumption (Wh) across datasets of varying scales, measured over the full end-to-end method pipeline. Power values correspond to the average power computed over total energy and runtime.}\label{tab:power_energy_tsc}
\centering
\setlength{\tabcolsep}{5pt} % Tightened padding to fit 5 columns naturally
\renewcommand{\arraystretch}{1.1}
\begin{tabular}{lrrrr}
\toprule
 & \multicolumn{2}{c}{\textbf{HC2}} & \multicolumn{2}{c}{\textbf{MR-TabPFN}} \\
\cmidrule(lr){2-3} \cmidrule(lr){4-5}
\textbf{Dataset (Scale)} & \textbf{Rate (W)} & \textbf{Total (Wh)} & \textbf{Rate (W)} & \textbf{Total (Wh)} \\ 
\midrule
BeetleFly (S)      & 35.35 & 1.54   & 45.62  & 0.87   \\
ACSF1 (SM)         & 33.21 & 17.15  & 43.96  & 1.09   \\
CricketX (M)       & 34.51 & 28.33  & 61.87  & 10.09  \\
CinCECGTorso (ML)  & 33.07 & 102.76 & 52.42  & 1.66   \\
LSST (L)           & 36.80 & 144.68 & 102.61 & 34.90  \\ 
\bottomrule
\end{tabular}
\end{table}

\subsection{Extrinsic Regression}

To evaluate the versatility of our framework in predicting continuous targets, we extend our analysis to TSER. Following the experimental protocol established in Section~\ref{subsec:tsc}, we first conducted an internal validation to determine which feature extraction technique (catch22, MultiROCKET, or tsfresh) best complements the TabPFN 2.5 architecture for regression. We assess the relative performance of these variants using a CD diagram based on their average Root Mean Squared Error (RMSE) across the TSER benchmark.

\begin{figure}[!htp]
\centering
\includegraphics[width=0.8\linewidth]{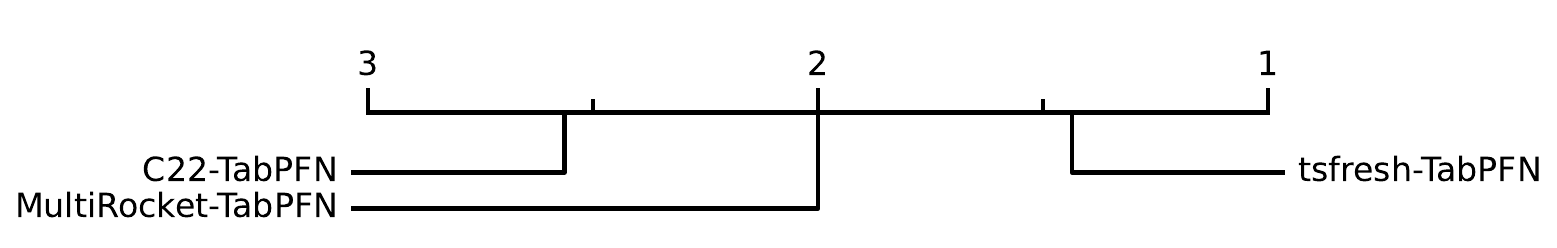}
\caption{Critical Difference diagram for TSER (Internal Validation) based on pairwise
Wilcoxon signed-rank tests of regression RMSE. Closer to 1, the better.}
\label{fig:tser_intra}
\end{figure}

The resulting CD diagram, illustrated in Figure \ref{fig:tser_intra}, reveals that the tsfresh-based configuration achieves the superior average rank (1.4364), with a statistically significant advantage over both competitors within this pipeline. This performance gap highlights the varying suitability of feature extraction strategies across different time series tasks. Specifically, while catch22 offers high computational speed, it consistently lacks the representational depth required for TabPFN to capture highly nuanced variances, ranking as the least effective feature extraction method across both TSC and TSER benchmarks.

Furthermore, while MultiROCKET's random kernels were highly effective for discerning class boundaries in classification, our results suggest that the exhaustive statistical descriptors generated by tsfresh are superior for regression. These comprehensive descriptors capture global characteristics and distribution shifts, enabling TabPFN's prior-fitted transformer to map input data more effectively to a continuous target space. This mapping appears more critical for extrinsic regression than the local pattern matching typical of kernel-based approaches, thereby justifying the selection of tsfresh as our primary extractor for this domain.

\begin{figure}[!htp]
\centering
\includegraphics[width=0.8\linewidth]{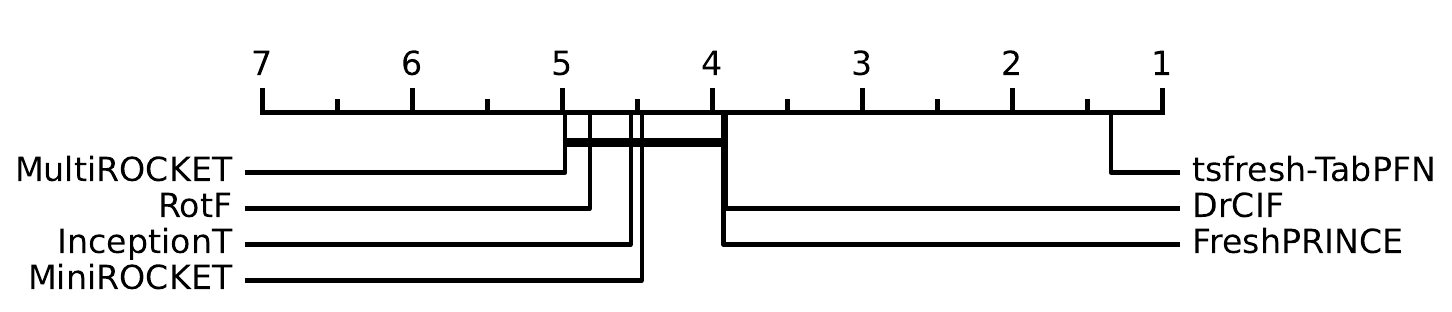}
\caption{Critical Difference for TSER (Comparative Analysis) based on pairwise Wilcoxon signed-rank tests of regression RMSE. Horizontal black bar indicates groups of models with no statistically significant difference. Closer to 1, the better.}
\label{fig:tser_bench}
\end{figure}

Upon identifying the optimal configuration (tsfresh-TabPFN), we benchmarked it against seven established TSER baselines, including specialized deep learning architectures (ResNet, InceptionTime) and robust feature-based methods (DrCIF, FreshPRINCE). As illustrated in the CD diagram in Figure \ref{fig:tser_bench}, tsfresh-TabPFN emerges as the top-performing model with a superior average rank of 1.4364. The absence of a horizontal bar connecting tsfresh-TabPFN to state-of-the-art models, such as DrCIF and FreshPRINCE, confirms a statistically significant improvement in predictive performance across the benchmark suite ($p < 0.05$). While methods like FreshPRINCE also utilize tsfresh features, the integration with TabPFN's prior-fitted transformer allows our framework to achieve a more precise mapping of continuous targets. These results represent an advancement over existing TSER methodologies, combining the descriptive power of exhaustive feature extraction with the predictive strength of tabular foundation models.

\begin{figure}[!htp]
    \centering
    \includegraphics[width=1\linewidth]{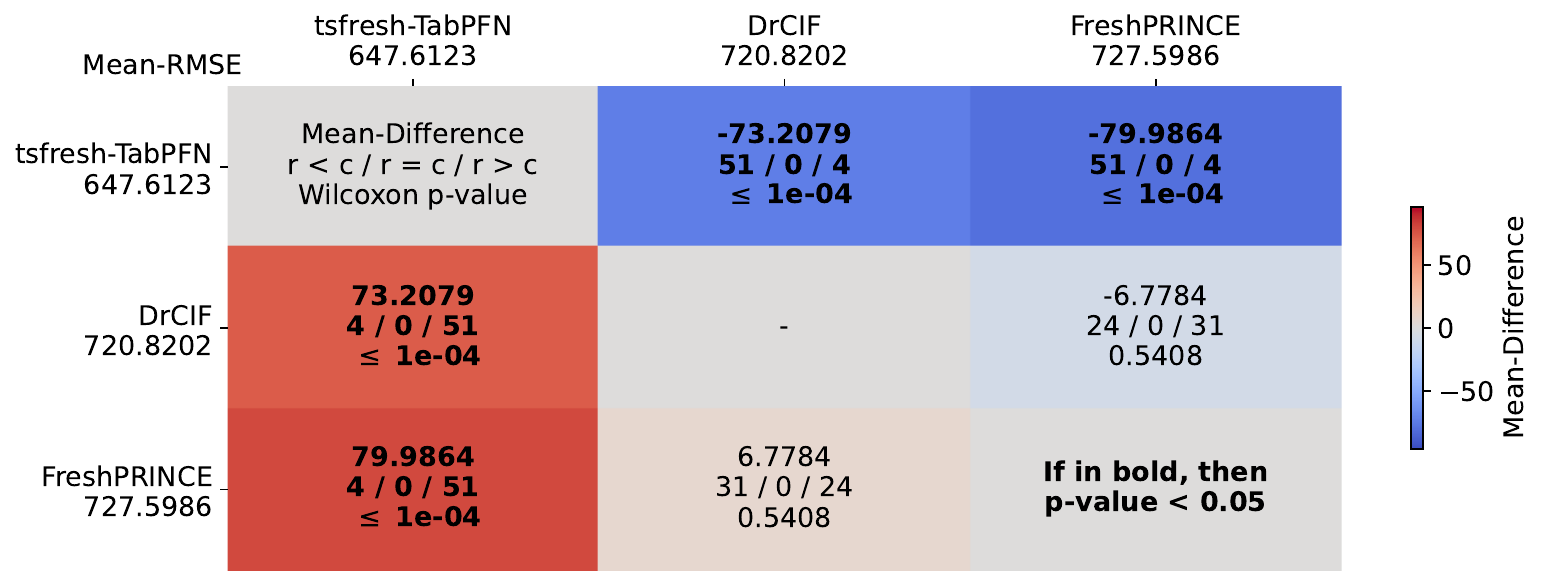}
    \caption{Multiple comparison matrix showing mean differences, win/tie/loss records and
Wilcoxon p-values. Bold values indicate statistical significance ($p < 0.05$).}
    \label{fig:tser_mcm}
\end{figure}

The pairwise comparison matrix, illustrated in Figure~\ref{fig:tser_mcm}, confirms tsfresh-TabPFN as the top-performing model. It demonstrates clear statistical dominance ($p \le 10^{-4}$) over state-of-the-art methods, outperforming both DrCIF and FreshPRINCE with identical 51/0/4 win records. Ultimately, tsfresh-TabPFN establishes a robust new benchmark for the TSER task.

The computational advantages of TS2TabPFN, previously observed in classification tasks, extend to TSER. While TSC experiments against HC2 reported speedups of up to 98.39×, TSER results against the DrCIF ensemble (Table~\ref{tab:runtime_tser}) reach up to 101.81×. These gains stem from replacing the expensive model fitting procedures of interval-based ensembles with a single forward-pass inference conditioned on extracted features. This is most pronounced in the BIDMC32SpO2 dataset, where DrCIF required over 48 hours ($\approx$196{,}498 s) compared to approximately 4 hours ($\approx$14{,}463 s) for tsfresh-TabPFN. Overall, TS2TabPFN achieves runtime reductions by decoupling feature extraction from a frozen foundation model, enabling a more scalable, training-free pipeline for large-scale settings.

\begin{table}[!htb]
\caption{Computational runtime and speedup factor for the tsfresh-TabPFN framework compared to the DrCIF baseline across datasets of varying scales.}\label{tab:runtime_tser}
\centering
\setlength{\tabcolsep}{10pt} % Espaçamento generoso entre colunas
\renewcommand{\arraystretch}{1.1} % Respiro vertical
\begin{tabular}{lrrr} % Esquerda para nomes, Direita para números
\toprule
\textbf{Dataset (Scale)} & \textbf{DrCIF (s)} & \textbf{tsfresh-TabPFN (s)} & \textbf{Speedup} \\ 
\midrule
Covid3Month (S)      &  194.01    &  4.04      &  48.02$\times$  \\
FloodModeling3 (SM)  & 1,123.95   &  11.04     & 101.81$\times$  \\
HouseholdPC1 (M)     & 28,688.97  &  943.90    &  30.39$\times$  \\
IEEEPPG (ML)         & 37,811.27  & 2,946.47   &  12.83$\times$  \\
BIDMC32SpO2 (L)      & 196,497.98 & 14,463.42  &  13.59$\times$  \\ 
\bottomrule
\end{tabular}
\end{table}

When isolating inference time alone, as reported in Table~\ref{tab:inference_tser}, tsfresh-TabPFN consistently outperforms DrCIF across all datasets, with speedups ranging from $12.61\times$ to $769.20\times$. This gap becomes more pronounced on larger datasets, reflecting the cost of repeated tree-based traversal and feature aggregation in ensemble methods, in contrast to the TabPFN-based model's single forward-pass inference conditioned on precomputed time series features.

\begin{table}[!htb]
\caption{Inference runtime and speedup factor comparison between tsfresh-TabPFN and DrCIF baseline during testing across representative datasets of varying scales.}\label{tab:inference_tser}
\centering
\setlength{\tabcolsep}{8pt}
\renewcommand{\arraystretch}{1.1}
\begin{tabular}{lrrr}
\toprule
\textbf{Dataset} & \textbf{DrCIF (s)} & \textbf{tsfresh-TabPFN (s)} & \textbf{Speedup} \\ 
\midrule
Covid3Month (S)             & 41.49     & 3.29   & 12.61$\times$ \\
FloodModeling3 (SM)         & 217.16    & 4.69   & 46.30$\times$ \\
HouseholdPC1 (M)            & 7,616.42  & 9.99   & 762.40$\times$ \\
IEEEPPG (ML)                & 9,225.97  & 21.28  & 433.55$\times$ \\
BIDMC32SpO2 (L)             & 58,789.46 & 76.43  & 769.20$\times$ \\ 
\bottomrule
\end{tabular}
\end{table}

Finally, Table~\ref{tab:power_energy_tser} reports hardware power consumption and total energy usage across datasets. While tsfresh-TabPFN exhibits higher power due to GPU utilization, its substantially shorter execution time results in significantly lower total energy consumption. For instance, on BIDMC32SpO2, energy usage decreases from 2,180.46 Wh to 456.13 Wh, confirming improved efficiency at scale.

\begin{table}[!htp]
\caption{Hardware power rate (W) and cumulative energy consumption (Wh) across datasets of varying scales, measured over the full end-to-end method pipeline. Power values correspond to the average power computed over total energy and runtime.}\label{tab:power_energy_tser}
\centering
\setlength{\tabcolsep}{5pt}
\renewcommand{\arraystretch}{1.1}
\begin{tabular}{lrrrr}
\toprule
 & \multicolumn{2}{c}{\textbf{DrCIF}} & \multicolumn{2}{c}{\textbf{tsfresh-TabPFN}} \\
\cmidrule(lr){2-3} \cmidrule(lr){4-5}
\textbf{Dataset} & \textbf{Rate (W)} & \textbf{Total (Wh)} & \textbf{Rate (W)} & \textbf{Total (Wh)} \\ 
\midrule
Covid3Month (S)            & 39.66 & 1.74    & 60.52  & 0.16 \\
FloodModeling3 (SM)        & 39.71 & 7.89    & 93.36  & 0.41 \\
HouseholdPC1 (M)           & 39.88 & 176.33  & 96.22  & 59.41 \\
IEEEPPG (ML)               & 39.85 & 234.93  & 96.10  & 90.66 \\
BIDMC32SpO2 (L)            & 39.95 & 2,180.46 & 96.27  & 456.13 \\ 
\bottomrule
\end{tabular}
\end{table}

\section{Limitations}
\label{sec:limitations}

Despite its strong empirical performance, the TS2TabPFN framework inherits specific constraints from its underlying architectural components. These limitations primarily pertain to memory scalability during both the feature extraction and the in-context inference phases. First, while the high-dimensional spaces of tsfresh and MultiROCKET can increase memory footprints on large datasets, this is easily mitigated by extracting features in batches, which preserves feature integrity and ensures compatibility with limited system RAM.

Second, the inference phase poses hardware constraints linked to the model's design. The ICL nature of TabPFN requires loading the entire dataset context into GPU memory simultaneously for a single forward pass. This results in memory usage that scales quadratically with context size due to the self-attention mechanism. Moreover, these physical constraints are compounded by TabPFN 2.5's internal limits when handling high-dimensional inputs. When a dataset exceeds the 500-feature threshold, the model relies on an ensemble-based feature subsampling strategy, increasing the number of required forward passes. Consequently, for high-dimensional spaces generated by tsfresh or MultiROCKET, this design introduces computational and memory overhead during inference.

\section{Conclusion}

The proposed TS2TabPFN framework introduces an effective approach to time series analysis by bridging specialized temporal feature extraction with tabular foundation models. By leveraging ICL, TS2TabPFN achieves strong predictive performance, matching elite ensembles such as HC2 in classification and establishing a new benchmark in extrinsic regression. Importantly, this performance is obtained while achieving speedups of up to two orders of magnitude and reducing processing times from days to hours on large-scale datasets. A dedicated inference-time analysis further confirms that these gains are not solely attributable to the training elimination inherent to ICL. In addition, an energy-efficiency evaluation using CodeCarbon shows that, despite a higher instantaneous power draw, the reduced execution time results in substantially lower total energy consumption compared to traditional ensemble pipelines.

Future work will explore richer temporal representations by integrating latent features from time series foundation models such as Mantis~\cite{feofanov2025mantis}. We also plan to investigate robustness under data scarcity and study how performance scales with reduced context windows, to extend the framework to massive datasets under strict memory constraints while preserving its training-free nature.

\begin{credits}

\subsubsection{\ackname}
The authors thank the maintainers of the UCR Time Series Classification Archive and the UEA \& UCR Time Series Extrinsic Regression Repository for making these valuable resources publicly available. This work was supported by grants \#2025/07753-1 and \#2022/03176-1, São Paulo Research Foundation (FAPESP).

\subsubsection{\discintname}
The authors have no competing interests to declare that are relevant to the content of this article.

\end{credits}

\bibliographystyle{splncs04}
\bibliography{mybibliography}

@article{tan2022multirocket,
    title={MultiRocket: multiple pooling operators and transformations for fast and effective time series classification: CW Tan},
    author={Tan, Chang Wei and Dempster, Angus and Bergmeir, Christoph and Webb, Geoffrey I},
    journal={Data Mining and Knowledge Discovery},
    volume={36},
    number={5},
    pages={1623--1646},
    year={2022},
    publisher={Springer}
}

@article{christ2018time,
    title={Time series feature extraction on basis of scalable hypothesis tests (tsfresh--a python package)},
    author={Christ, Maximilian and Braun, Nils and Neuffer, Julius and Kempa-Liehr, Andreas W},
    journal={Neurocomputing},
    volume={307},
    pages={72--77},
    year={2018},
    publisher={Elsevier}
}

@article{lubba2019catch22,
    title={catch22: CAnonical Time-series CHaracteristics: Selected through highly comparative time-series analysis},
    author={Lubba, Carl H and Sethi, Sarab S and Knaute, Philip and Schultz, Simon R and Fulcher, Ben D and Jones, Nick S},
    journal={Data mining and knowledge discovery},
    volume={33},
    number={6},
    pages={1821--1852},
    year={2019},
    publisher={Springer}
}

@article{dempster2020rocket,
    title={ROCKET: exceptionally fast and accurate time series classification using random convolutional kernels},
    author={Dempster, Angus and Petitjean, Fran{\c{c}}ois and Webb, Geoffrey I},
    journal={Data Mining and Knowledge Discovery},
    volume={34},
    number={5},
    pages={1454--1495},
    year={2020},
    publisher={Springer}
}

@article{ismail2020inceptiontime,
    title={Inceptiontime: Finding alexnet for time series classification},
    author={Ismail Fawaz, Hassan and Lucas, Benjamin and Forestier, Germain and Pelletier, Charlotte and Schmidt, Daniel F and Weber, Jonathan and Webb, Geoffrey I and Idoumghar, Lhassane and Muller, Pierre-Alain and Petitjean, Fran{\c{c}}ois},
    journal={Data mining and knowledge discovery},
    volume={34},
    number={6},
    pages={1936--1962},
    year={2020},
    publisher={Springer}
}

@article{middlehurst2024bake,
    title={Bake off redux: a review and experimental evaluation of recent time series classification algorithms: M. Middlehurst et al.},
    author={Middlehurst, Matthew and Sch{\"a}fer, Patrick and Bagnall, Anthony},
    journal={Data Mining and Knowledge Discovery},
    volume={38},
    number={4},
    pages={1958--2031},
    year={2024},
    publisher={Springer}
}

@article{mohammadi2024deep,
    title={Deep learning for time series classification and extrinsic regression: A current survey},
    author={Mohammadi Foumani, Navid and Miller, Lynn and Tan, Chang Wei and Webb, Geoffrey I and Forestier, Germain and Salehi, Mahsa},
    journal={ACM Computing Surveys},
    volume={56},
    number={9},
    pages={1--45},
    year={2024},
    publisher={ACM New York, NY}
}

@inproceedings{middlehurst2022freshprince,
    title={The freshprince: A simple transformation based pipeline time series classifier},
    author={Middlehurst, Matthew and Bagnall, Anthony},
    booktitle={International Conference on Pattern Recognition and Artificial Intelligence},
    pages={150--161},
    year={2022},
    organization={Springer}
}

@article{guijo2024unsupervised,
    title={Unsupervised feature based algorithms for time series extrinsic regression},
    author={Guijo-Rubio, David and Middlehurst, Matthew and Arcencio, Guilherme and Silva, Diego Furtado and Bagnall, Anthony},
    journal={Data Mining and Knowledge Discovery},
    volume={38},
    number={4},
    pages={2141--2185},
    year={2024},
    publisher={Springer}
}

@article{bagnall2018rotation,
    title={Is rotation forest the best classifier for problems with continuous features?},
    author={Bagnall, Anthony and Flynn, M and Large, J and Line, J and Bostrom, A and Cawley, G},
    journal={arXiv preprint arXiv:1809.06705},
    year={2018}
}

@inproceedings{wang2017time,
    title={Time series classification from scratch with deep neural networks: A strong baseline},
    author={Wang, Zhiguang and Yan, Weizhong and Oates, Tim},
    booktitle={2017 International joint conference on neural networks (IJCNN)},
    pages={1578--1585},
    year={2017},
    organization={IEEE}
}

@misc{tsml_archive_maintainers_2024_11236865,
    author = {{TSML Archive Maintainers}},
    title  = {TSML ƒExtended Time Series Extrinsic Regression Archive 2024},
    month  = may,
    year = 2024,
    publisher = {Zenodo},
    doi = {10.5281/zenodo.11236865}
}

@misc{UCRArchive2018,
    title = {The {UCR} Time Series Classification Archive},
    author = {Dau, Hoang Anh and Keogh, Eamonn and Kamgar, Kaveh and Yeh, Chin-Chia Michael and Zhu, Yan and Gharghabi, Shaghayegh and Ratanamahatana, Chotirat Ann and Yanping and Hu, Bing and Begum, Nurjahan and Bagnall, Anthony and Mueen, bdullah and Batista, Gustavo
    and {Hexagon-ML}},
    year = {2018},
    month = {October},
}

@misc{bagnall2018ueamultivariatetimeseries,
    title={The UEA multivariate time series classification archive, 2018},
    author={Anthony Bagnall and Hoang Anh Dau and Jason Lines and Michael Flynn and James Large and Aaron Bostrom and Paul Southam and Eamonn Keogh},
    year={2018},
    eprint={1811.00075},
    archivePrefix={arXiv},
    primaryClass={cs.LG},
}

@article{aeon24jmlr,
    author = {Matthew Middlehurst and Ali Ismail-Fawaz and Antoine Guillaume and Christopher Holder and David Guijo-Rubio and Guzal Bulatova and Leonidas Tsaprounis and Lukasz Mentel and Martin Walter and Patrick Sch{{\"a}}fer and Anthony Bagnall},
    title  = {aeon: a Python Toolkit for Learning from Time Series},
    journal = {Journal of Machine Learning Research},
    year = {2024},
    volume = {25},
    number = {289},
    pages = {1--10},
}

@article{demvsar2006statistical,
    title={Statistical comparisons of classifiers over multiple data sets},
    author={Dem{\v{s}}ar, Janez},
    journal={Journal of Machine learning research},
    volume={7},
    number={Jan},
    pages={1--30},
    year={2006}
}

@inproceedings{hollmann2023tabpfn,
    title={TabPFN: A Transformer That Solves Small Tabular Classification Problems in a Second},
    author={Hollmann, Noah and M{\"u}ller, Samuel and Eggensperger, Katharina and Hutter, Frank},
    booktitle={The Eleventh International Conference on Learning Representations},
    year={2023}
}

@article{middlehurst2021hive,
    title={HIVE-COTE 2.0: a new meta ensemble for time series classification},
    author={Middlehurst, Matthew and Large, James and Flynn, Michael and Lines, Jason and Bostrom, Aaron and Bagnall, Anthony},
    journal={Machine Learning},
    volume={110},
    number={11},
    pages={3211--3243},
    year={2021},
    publisher={Springer}
}

@article{grinsztajn2025tabpfn,
    title={Tabpfn-2.5: Advancing the state of the art in tabular foundation models},
    author={Grinsztajn, L{\'e}o and Fl{\"o}ge, Klemens and Key, Oscar and Birkel, Felix and Jund, Philipp and Roof, Brendan and J{\"a}ger, Benjamin and Safaric, Dominik and Alessi, Simone and Hayler, Adrian and others},
    journal={arXiv preprint arXiv:2511.08667},
    year={2025}
}

@article{feofanov2025mantis,
    title={Mantis: Lightweight calibrated foundation model for user-friendly time series classification},
    author={Feofanov, Vasilii and Wen, Songkang and Alonso, Marius and Ilbert, Romain and Guo, Hongbo and Tiomoko, Malik and Pan, Lujia and Zhang, Jianfeng and Redko, Ievgen},
    journal={arXiv preprint arXiv:2502.15637},
    year={2025}
}

@article{tan2021time,
    title={Time series extrinsic regression: Predicting numeric values from time series data},
    author={Tan, Chang Wei and Bergmeir, Christoph and Petitjean, Fran{\c{c}}ois and Webb, Geoffrey I},
    journal={Data Mining and Knowledge Discovery},
    volume={35},
    number={3},
    pages={1032--1060},
    year={2021},
    publisher={Springer}
}

@article{dau2019ucr,
    title={The UCR time series archive},
    author={Dau, Hoang Anh and Bagnall, Anthony and Kamgar, Kaveh and Yeh, Chin-Chia Michael and Zhu, Yan and Gharghabi, Shaghayegh and Ratanamahatana, Chotirat Ann and Keogh, Eamonn},
    journal={IEEE/CAA Journal of Automatica Sinica},
    volume={6},
    number={6},
    pages={1293--1305},
    year={2019},
    publisher={IEEE}
}

@article{lacoste2019quantifying,
    title={Quantifying the Carbon Emissions of Machine Learning},
    author={Lacoste, Alexandre and Luccioni, Alexandra and Schmidt, Victor and Dandres, Thomas},
    journal={arXiv preprint arXiv:1910.09700},
    year={2019}
}

\end{document}